\documentclass{article}

 \usepackage[preprint]{neurips_2025}

\usepackage{enumitem}    
\usepackage{setspace}    
\usepackage[utf8]{inputenc} 
\usepackage[T1]{fontenc}    
\usepackage{url}            
\usepackage{booktabs}       
\usepackage{amsfonts}       
\usepackage{nicefrac}       
\usepackage{microtype}      
\usepackage{xcolor}         
\usepackage{amsmath}        

\usepackage[ruled,vlined,linesnumbered]{algorithm2e}
\DontPrintSemicolon
\SetKwInput{KwIn}{Input}
\SetKwInput{KwOut}{Output}

\usepackage{graphicx}
\usepackage{float}

\usepackage[colorlinks=true, linkcolor=blue, citecolor=blue, urlcolor=blue]{hyperref}
\usepackage{bookmark}
\usepackage{cleveref}

\setlist[itemize]{
  leftmargin=*,
  itemsep=0pt,
  topsep=0pt,
  partopsep=0pt,
  parsep=0pt
}

\title{Opinion: Towards Unified Expressive Policy Optimization for Robust Robot Learning}

\author{%
Haidong Huang$^{1,2*}$ \quad
Haiyue Zhu$^{3*}$ \quad
Jiayu Song$^{1}$ \quad
Xixin Zhao$^{1}$ \quad
Yaohua Zhou$^{1}$ \\
\textbf{Jiayi Zhang}$^{2}$ \quad
\textbf{Yuze Zhai}$^{4}$ \quad
\textbf{Xiaocong Li}$^{1\dagger}$ \\
\\
$^{1}$ Eastern Institute of Technology, Ningbo \\
$^{2}$ University of Nottingham \\
$^{3}$ SIMTech, Agency for Science, Technology and Research (A*STAR) \\
$^{4}$ Southern University of Science and Technology \\
\\
$^{*}$ Equal contribution \quad $^{\dagger}$ Corresponding author
}

\begin{document}

\maketitle

\begin{abstract}
Offline-to-online reinforcement learning (O2O-RL) has emerged as a promising paradigm for safe and efficient robotic policy deployment but suffers from two fundamental challenges: limited coverage of multimodal behaviors and distributional shifts during online adaptation. We propose UEPO, a unified generative framework inspired by large language model pretraining and fine-tuning strategies. Our contributions are threefold: (1) a multi-seed dynamics-aware diffusion policy that efficiently captures diverse modalities without training multiple models; (2) a dynamic divergence regularization mechanism that enforces physically meaningful policy diversity; and (3) a diffusion-based data augmentation module that enhances dynamics model generalization. On the D4RL benchmark, UEPO achieves +5.9\% absolute improvement over Uni-O4 on locomotion tasks and +12.4\% on dexterous manipulation, demonstrating strong generalization and scalability.
\end{abstract}

\section{INTRODUCTION} 

Offline-to-Online Reinforcement Learning (O2O-RL) has emerged as a key paradigm for safe and efficient robot deployment. It leverages static offline datasets to pretrain base policies so as to more accurately capture physical dynamics, mitigating real-world trial-and-error risks, and further fine-tunes policies via environmental interaction to adapt to dynamic scenarios, forming a complete “offline initialization to online fine-tuning” framework. In practice, the perception stack and camera configuration materially affect data quality in the wild: camera motion can confound evaluation, and large camera arrays require stable calibration and efficient refinement \citep{lin2025camerabench, you2025multi, Li_2024_BMVC}. In adjacent agentic systems, LLM-based multi-agent orchestration and self-evolving teamwork demonstrate scalable planning capabilities \citep{chen2025mdteamgptselfevolvingllmbasedmultiagent}, while systematic safety audits reveal risk accumulation in multi-agent decision loops and motivate conservative online interaction \citep{chen2025medsentryunderstandingmitigatingsafety}; furthermore, learnable tool capability memory helps agents choose and parameterize tools more reliably \citep{xiao2025toolmem}. However, existing O2O-RL methods still face significant challenges, including inefficient offline initialization and a weak interface between generative models and online adaptation \citep{zhang2024samg,laria2024assessing}.

Traditional Behavior Cloning (BC) \citep{bai2025rethinking} relies heavily on large amounts of expert data and struggles to cover multi-modal action distributions. While mainstream generative models such as Diffusion Policy \citep{chi2024diffusionpolicy} excel at offline modeling, their fixed noise schedules and the lack of environmental feedback often lead to policy degradation and distribution shift during online fine-tuning \citep{ma2025efficient}.

Recent frameworks such as Off2On \citep{hong2022confidence}, BPPO \citep{zhuang2023behavior}, and Uni-O4 \citep{lei2024unio} have made notable progress in integrating offline and online learning by jointly optimizing objectives without additional regularization. Nevertheless, Uni-O4 \citep{lei2024unio} still exhibits limitations in offline pre-training, generative model adaptation, and scalability, leading to high computational costs for the integrated strategy, insufficient diversity at the physical execution level, and poor data efficiency and generalization. Moreover, current pipelines rarely leverage advances in upstream data selection and preference aggregation that are commonplace in multimodal recommendation, such as multi-level self-supervised alignment for retrieval and group-consensus modeling \citep{xu2025mentor,xu2024aligngroup}. These shortcomings render existing algorithms less suitable for high-dimensional dynamics and scenarios with scarce real-world data.

To address these issues, we propose a unified generative O2O-RL framework. Our approach includes a data-efficient generative offline module that: (1) employs a dynamics-aware diffusion policy combining U-Net \citep{ronneberger2015u} and Transformer \citep{vaswani2017attention} to model long-horizon action sequences, while using different noise seeds to generate diverse sub-policies without training multiple models, significantly reducing the need for expert demonstrations; (2) integrates divergence regularization with diffusion sampling diversity to enhance behavioral differences among sub-policies through dynamic discrepancy measurement, noise perturbation, and sequence-level constraints; (3) expands training data by synthesizing trajectories through diffusion and then combines them with real data to train dynamics models. This effectively bridges offline generative strategies with online fine-tuning.

Our framework enhances policy representation, diversity, and generalization, improving O2O-RL for complex robotic tasks. On the D4RL benchmark, it surpasses state-of-the-art baselines, especially in dexterous manipulation and quadruped locomotion, showing strong stability and adaptability.

\vspace{-4mm} 

\section{METHOD}
\begin{figure}[h!]
    \centering
    \includegraphics[width=\linewidth]{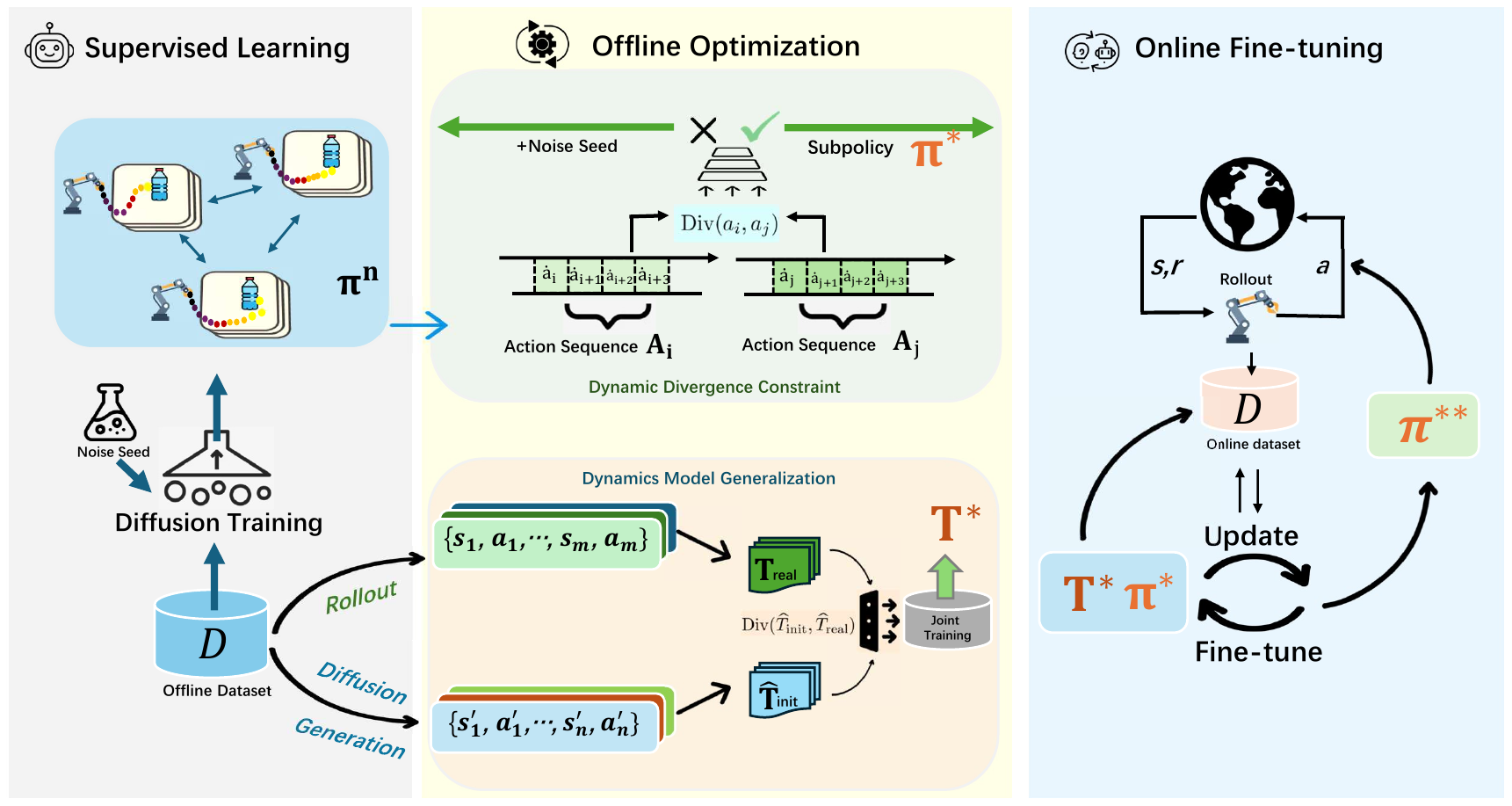}
    \caption{UEPO employs a multi-seed diffusion sampling strategy to initialize components for the subsequent phase. During the offline optimization stage (middle), the strategy enhances diversity through a regularization mechanism that amplifies policy divergence, whilst simultaneously training the dynamics model $\hat{T}$ using a joint approach of real data and synthetic trajectories to improve its generalization capability. Finally, a qualifying policy is selected as the initialization for online fine-tuning.}
    \label{dynamics} 
\end{figure}

In this section, we present our method, UEPO, as shown in Fig. \ref{dynamics}. The three core innovations of this framework are seamlessly integrated into the learning pipeline spanning the offline and online phases.

\vspace{-4mm} 

\subsection{Conditional Action Sequence Generation via Diffusion Model}
\label{subsec:diffusion_policy}
To address limitations of BC, we adopt a state-conditional diffusion policy to model the entire action sequence distribution $p(a_{1:T} \mid s_{1:T})$, capturing long-horizon dependencies and multi-modal behaviors in offline data. We then construct an ensemble policy via multi-seed sampling to enhance sequence rationality and behavioral diversity.

Traditional ensemble methods incur high computational costs due to the training of multiple independent models. Instead, we construct an ensemble of $n$ sub-policies $\{\pi_\theta^i\}_{i=1}^n$ from a single trained diffusion model by varying initial noise seeds during reverse sampling. For each sub-policy $\pi_\theta^i$, we condition on the same state sequence $s_{1:T}$ but initialize the reverse process with a distinct random seed $\epsilon_i \sim \mathcal{N}(0, \mathbf{I})$. Each unique seed generates an action sequence $a_{1:T}^i$ corresponding to a distinct behavioral modality, thus reducing training costs while naturally promoting sub-policy diversity.

\subsection{Divergence Regularization Enhancement Guided by Diffusion Sampling}
\label{subsec:divergence}
The ensemble policy constructed in Section~\ref{subsec:diffusion_policy} provides initial diversity. However, to ensure sub-policies exhibit divergence during dynamic execution, we introduce dynamic divergence constraints directly into the diffusion sampling process. This contrasts with Uni-O4's (\citet{lei2024unio}) approach, which applies a KL divergence penalty to the distribution of single-step actions and may result in  policies that are statically distinct but dynamically similar or mutually conflicting.

\subsubsection{Dynamic Divergence Constraint}
When generating action sequence $a_i$ for the $i$-th sub-policy, we measure its divergence from other sub-policies $\{a_j \mid j < i\}$ that have already been generated within the same sampling round. We introduce a \textit{divergence reward} based on dynamics-level discrepancies to adjust the sampling path.
\begin{itemize}
    \item \textbf{Dynamic Divergence Metric:} We define the divergence between two action sequences $a_i$ and $a_j$ by measuring the dynamics difference between first-order (velocity) and second-order (acceleration):
    \[
    \text{div}(a_i, a_j) = \frac{1}{T} \sum_{t=1}^{T} \left( \| \dot{a}_{i,t} - \dot{a}_{j,t} \|_2 + (1 - \cos(\ddot{a}_{i,t}, \ddot{a}_{j,t})) \right)
    \]
    where velocity $\dot{a}_t = a_t - a_{t-1}$ and acceleration $\ddot{a}_t = \dot{a}_t - \dot{a}_{t-1}$, to ensure that discrepancies are meaningful at the level of physical execution.
    \item \textbf{Adaptive Perturbation:} If the divergence $\text{div}(a_i, a_j)$ falls below a threshold $\tau$, we interpret the paths as being too similar. To encourage exploration, we perturb the current denoised estimate $a_t^i$ in the reverse process:
    \[
    a_t^i \leftarrow a_t^i + \delta, \quad \delta \sim \mathcal{N}(0, \sigma_{\text{div}}^2 \mathbf{I}), \quad \text{where } \sigma_{\text{div}} = \eta \cdot \frac{\tau - \text{div}(a_i, a_j)}{\tau}
    \]
    The scaling factor $\sigma_{\text{div}}$ increases as the divergence decreases, and $\eta$ is a hyperparameter controlling the perturbation strength. This adaptive noise injection forces the sub-policy to explore distinct dynamic modes.
\end{itemize}

\subsubsection{Synergy with Sequence-Level KL Regularization}
We retain the KL divergence penalty from Uni-O4 (\citet{lei2024unio}) to ensure distributional diversity at a global level. However, we redefine its application  from the single-step action distribution to the entire action-sequence distribution, which aligns naturally with our sequence-based diffusion policy.

The overall objective for each sub-policy $\hat{\pi}^i$ is:
\[
J(\hat{\pi}^i) = \mathbb{E}_{(s, a) \sim \mathcal{D}} \left[ \log p_\theta(a \mid s) \right] + \alpha \, \mathbb{E}_{(s, a) \sim \mathcal{D}} \left[ \log \left( \frac{p_\theta(a \mid s)}{\max_{j} p_\theta(a \mid s)} \right) \right]
\]
Here, $p_\theta(a \mid s)$ represents the probability of generating the entire action sequence $a$ given state $s$, approximated by the product of Markov transition probabilities in the reverse diffusion process. This combination of a local dynamic constraint and a global sequence-level regularizer effectively enhances sub-policy diversity.

\subsection{Enhancing Dynamics Model Generalization with Diffusion}
\label{subsec:dynamics}
A common challenge in model-based RL (\citet{jiang2016doubly}) is the limited generalization of the
learned dynamics model $\hat{T}(s' \mid s, a)$ when the offline data set $\mathcal{D}$ does not adequately 
cover the state-action space. To mitigate Uni-O4 (\citet{lei2024unio})'s potential overfitting to limited 
transitions, we use our diffusion policy to generate physically plausible  trajectories for 
augmenting the training data of the dynamics model, thereby enhancing its generalization and the 
sample efficiency of online learning. 
This procedure generates virtual trajectories consistent with real dynamics, 
as shown in Algorithm~\ref{alg:trajectory}, thereby providing reliable augmented data for joint model training.

\vspace{-1mm}
\begin{algorithm}[htbp]
\caption{Virtual Trajectory Generation and Filtering}
\label{alg:trajectory}
\KwIn{Initial state $s_0 \sim \rho_{\mathcal{D}}$; pre-trained diffusion policy $\pi_{\text{diff}}$; 
real transition dynamics $T_{\text{real}}$; initial dynamics model $\hat{T}_{\text{init}}$; 
threshold $\epsilon=0.05$}
\KwOut{Filtered trajectory dataset $\mathcal{D}_{\text{diff}}$}

Initialize $\mathcal{D}_{\text{diff}} \gets \emptyset$\;
Generate multi-step action sequence $a_{0:T-1}$ using $\pi_{\text{diff}}$ conditioned on $s_0$\;
Initialize trajectory $\tau \gets \emptyset$\;

\For{$t=0$ \KwTo $T-1$}{
  Sample $s_{t+1} \sim T_{\text{real}}(\cdot \mid s_t, a_t)$\;
  Add $(s_t, a_t, s_{t+1})$ to $\tau$\;
}

Trajectory $\tau = \{(s_0,a_0,s_1),(s_1,a_1,s_2),\dots,(s_{T-1},a_{T-1},s_T)\}$\;

Compute $D_{\mathrm{KL}}(T_{\text{real}}(s'|s,a)\,\|\, \hat{T}_{\text{init}}(s'|s,a))$\;

\If{$D_{\mathrm{KL}} < \epsilon$ \tcp*[r]{Filtering criterion}}{
  Add $\tau$ to $\mathcal{D}_{\text{diff}}$\;
}

\tcp{Ensures augmented data remains consistent with the underlying physics}
\Return{$\mathcal{D}_{\text{diff}}$}
\end{algorithm}

\vspace{2mm}

\textbf{Joint Training of Dynamics Models}

The original maximum likelihood objective of the dynamics model is updated to incorporate the filtered virtual trajectories $\mathcal{D}_{\text{diff}}$, creating a joint training dataset:
\[
\mathcal{L}(\hat{T}) = -\mathbb{E}_{(s,a,s') \sim \mathcal{D} \cup \mathcal{D}_{\text{diff}}} 
\left[ \log \hat{T}(s' \mid s,a) \right]
\]

The size of $\mathcal{D}_{\text{diff}}$ is controlled to be 2–3 times that of $\mathcal{D}$ ($|\mathcal{D}_{\text{diff}}| \approx 2|\mathcal{D}|$), striking a balance between augmenting data volume and maintaining the fidelity of the real data distribution. This process significantly improves the model’s ability to generalize to unseen regions of the state–action space.

\section{EXPERIMENTS}

We evaluate our proposed algorithm across extensive offline RL benchmarks to investigate two aspects: (i) its performance compared to state-of-the-art baselines and (ii) its capability to model the multimodal nature of offline datasets by integrating multiple sub-diffusion policies.

\begin{table}[H]  
\centering
\tiny  

\begin{tabular}{lcccccccccc} 
\toprule   
\textbf{Environment} & \textbf{CQL} & \textbf{TD3+BC} & \textbf{Onestep RL} & \textbf{IQL} & \textbf{COMBO} & \textbf{BPPO} & \textbf{ATAC} & \textbf{BC} & \textbf{UNI-O4}  & \textbf{Ours} \\
\midrule  
halfcheetah-medium-v2       & 44.0 & 48.3 & 48.4 & 47.4 & 54.2 & 44.0 & \textbf{54.3} & 42.1 & 52.6 & \textbf{57$\pm$0.8} \\
hopper-medium-v2            & 58.5 & 59.3 & 59.6 & 66.3 & 97.2 & 93.9 & 102.8 & 52.8 & 104.4 & \textbf{108$\pm$0.5} \\
walker2d-medium-v2          & 72.5 & 83.7 & 81.8 & 78.3 & 81.9 & 83.6 & \textbf{91.0} & 74.0 & 90.2 & \textbf{91$\pm$1.4} \\
halfcheetah-medium-replay   & 45.5 & 44.6 & 38.1 & 44.2 & 55.1  & 41.0 & 49.5 & 34.9 & 44.3 & \textbf{58.2$\pm$0.7} \\
hopper-medium-replay        & 95.0 & 60.9 & 97.5 & 97.7 & 89.5 & 92.5 & 102.8 & 25.7 & 103.2 & \textbf{112.0$\pm$2.3} \\
walker2d-medium-replay      & 77.2 & 81.8 & 49.5 & 73.9 & 96.0 & 77.6 & 94.1 & 54.9 & 98.4 & \textbf{103.8$\pm$1.7} \\
halfcheetah-medium-expert   & 91.6 & 90.7 & 93.4 & 89.7 & 90.0 & 92.6 &\textbf{ 95.5} & 52.9 & 93.8 & 94.3$\pm$0.6 \\
hopper-medium-expert        & 105.4 & 98.0 & 103.3 & 91.7 & 111.1 & 112.8 & 112.6 & 18.6 & 111.4 & \textbf{118.6$\pm$0.2} \\
walker2d-medium-expert      & 108.8 & 110.1 & 113.0 & 109.6 & 103.3 & 113.1 & 116.3 & 107.7 & 118.1 & \textbf{120.7$\pm$0.3} \\
\midrule  
\textbf{locomotion total}   & 698.5 & 677.4 & 684.6 & 692.4 & 738.3 & 751.0 & 818.9 & 463.5 & 816.4 &\textbf{ 864.6$\pm$8.5} \\
\midrule  
pen-human                   & 37.5 & 8.4 & 90.7 & 71.5 & 41.3 &117.8 & 79.3 & 65.8 & 116.2* & \textbf{122.8$\pm$5.8} \\
hammer-human                & 4.4 & 2.0 & 0.2 & 1.4 & 9.6 & 14.9 & 6.7 & 2.6 & 247.1 & \textbf{30.2$\pm$3.3} \\ 
door-human                  & 9.9 & 0.5 & -0.1 & 4.3 & 5.2 & 25.8 & 8.7 & 4.3 & 17.3* & \textbf{29.3$\pm$0.7} \\
relocate-human              & 0.2 & -0.3 & 2.1 & 0.1 & 0.4 & \textbf{4.8} & 0.3 & 0.2 & 27.1* & 2.9$\pm$0.7 \\
pen-cloned                  & 39.2 & 41.5 & 60.0 & 37.3 & 24.6 & 110.8 & 73.9 & 60.7 & 101.4* & \textbf{118.4$\pm$12.4} \\
hammer-cloned               & 2.1 & 0.8 & 2.0 & 2.1 & 3.3 & 8.9 & 2.3 & 0.4 & 7.3* & \textbf{9.7$\pm$0.8} \\
door-cloned                 & -0.1 & -0.4 & -0.4 & -1.6 & 0.2 & 6.2 & 8.2 & 0.9 & 10.2* & \textbf{9.8$\pm$2.4} \\
relocate-cloned             & 0.4 & -0.3 & 0.1 & 0.2 & 0.7 & \textbf{1.9 }& 0.8 & 0.1 & 1.4* & 1.3$\pm$0.4 \\
\midrule  
\textbf{Adroit total}       & 93.6 & 52.2 & 155.2 & 118.1 & 84.2 &291.4 & 180.2 & 135.0 & 288.6 & \textbf{324.4$\pm$26.5} \\
\midrule  
kitchen-complete            & 43.8 & 0.0 & 2.0 & 62.5 & 3.5 & 91.5 & 2.0 & 68.3 & 93.6 & \textbf{102.6$\pm$3.6} \\
kitchen-partial             & 49.8 & 22.5 & 35.5 & 46.3 & 1.2 & 57.0 & 0.0 & 32.5 & \textbf{58.3} & \textbf{57.6$\pm$2.8} \\
kitchen-mixed               & 51.0 & 25.0 & 28.0 & 51.0 & 1.4 & 62.5 & 1.0 & 47.5 & 65.0 & \textbf{70.3$\pm$5.6} \\S
\textbf{kitchen total}      & 144.6 & 47.5 & 65.5 & 159.8 & 6.1 & 211.0 & 3.0 & 148.3 & 216.9 & \textbf{230.5$\pm$12.0} \\
\midrule  
\textbf{Total}              & 936.7 & 777.1 & 905.3 & 970.3 & 828.6 & 1253.4 & 1002.1 & 746.8 & 1322.0 &\textbf{ 1419.5$\pm$47.0} \\
\bottomrule  
\end{tabular}

\caption{Most of the results are extracted from the original papers, and *indicates that the results are reproduced by running the provided source code.}
\end{table}

\vspace{-4mm}

\section{CONCLUSION}

In this work, we propose UEPO, a unified generative offline-to-online reinforcement learning framework that effectively addresses key limitations in existing O2O-RL approaches. Our approach integrates a dynamics-aware diffusion policy for efficient offline initialization and a differential-enhanced regularization mechanism to enhance policy diversity. Our approach mitigates distribution shifts, reduces reliance on expert data, and demonstrates strong generalization capabilities across complex, high-dimensional tasks. Experiments on the D4RL benchmark reveal that UEPO achieves state-of-the-art performance.

\bibliographystyle{plainnat}  
\bibliography{reference} 

\medskip

\end{document}